\documentclass[runningheads]{llncs}
\usepackage{graphicx}
\begin{document}
\title{FAIRification of MLC data
}
%
%
\author{Ana Kostovska\inst{1}
\and
Jasmin Bogatinovski\inst{2}
\and
Andrej Treven \inst{1}
\and
Sa\v{s}o D\v{z}eroski \inst{1}
\and
Dragi Kocev \inst{1}
\and
Pan\v{c}e Panov \inst{1}
}
\authorrunning{A. Kostovska et al.}
%
\institute{Jo\v{z}ef Stefan Institute, Department of Knowledge Technologies, Ljubljana, Slovenia \and
Technical University Berlin, Berlin, 10587, Germany
}
\maketitle              
\begin{abstract}
\begin{sloppypar}
The multi-label classification (MLC) task has increasingly been receiving interest from the machine learning (ML) community, as evidenced by the growing number of papers and methods that appear in the literature. Hence, ensuring proper, correct, robust, and trustworthy benchmarking is of utmost importance for the further development of the field. We believe that this can be achieved by adhering to the recently emerged data management standards, such as the FAIR (Findable, Accessible, Interoperable, and Reusable) and TRUST (Transparency, Responsibility, User focus, Sustainability, and Technology) principles. To FAIRify the MLC datasets, we introduce an ontology-based online catalogue of MLC datasets that follow these principles. The catalogue extensively describes many MLC datasets with comprehensible meta-features, MLC-specific semantic descriptions, and different data provenance information. The MLC data catalogue is extensively described in our recent publication in Nature Scientific Reports, Kostovska et al. \cite{kostovska2022fair}, and available at: \url{http://semantichub.ijs.si/MLCdatasets}. In addition, we provide an ontology-based system for easy access and querying of performance/benchmark data obtained from a comprehensive MLC benchmark study. The system is available at: \url{http://semantichub.ijs.si/MLCbenchmark}.
\end{sloppypar}
\end{abstract}
\section{Introduction}
MLC is a machine learning task where the goal is to predict the subset of labels that are relevant for a given data example \cite{Madjarov2012,ZhangTPAMI,Herrera2016,Moyano2018}. We have witnessed the broad use of the MLC methods in diverse interdisciplinary applications ranging from areas in biology, bioinformatics, chemistry, medicine, video, audio, images, text, and the number of applications is constantly increasing \cite{Gibaja2014,bogatinovski2022ESWA1,bogatinovski2022IJIS}. With the growing interest for the MLC task in various application domains, numerous MLC methods and datasets have been published. However, the data produced from the MLC studies are scattered across heterogeneous data sources, which limits data interoperability. Thus, there is an urgent need to improve MLC data management and support the reuse of scholarly data.

Recently, different data management principles have been gaining much traction among various scientific stakeholders. The FAIR principles \cite{wilkinson2016fair} are a set of guiding principles that have been introduced to support and promote proper data management and stewardship. The TRUST principles \cite{lin2020trust} go a level higher by focusing on data repositories and guiding their design and development.

This paper introduces two, FAIR- and TRUST-compliant, MLC data systems. The first one is an MLC data catalogue. The catalogue extensively describes many MLC datasets with comprehensible meta-features and different data provenance information. The meta-features represent various measurable properties of the learning task \cite{vanschoren2018meta}. Describing the MLC datasets with meta-features that capture the properties of the MLC task can allow for joint cross-domain investigation of the different MLC applications. The accumulation of meta-knowledge of this kind also allows the study of the task itself and improves the generalization performance. The second one is system for MLC benchmark data. The system offers access to a large set of benchmark data obtained for a comprehensive, comparative study of MLC methods that evaluates 26 methods on 42 benchmark MLC datasets using 20 evaluation measures. This data provides an overview of the performance landscape of the MLC methods.

\section{MLC data catalogue and benchmark data system}
\begin{figure}[t]
\centering
\includegraphics[width=0.9\textwidth]{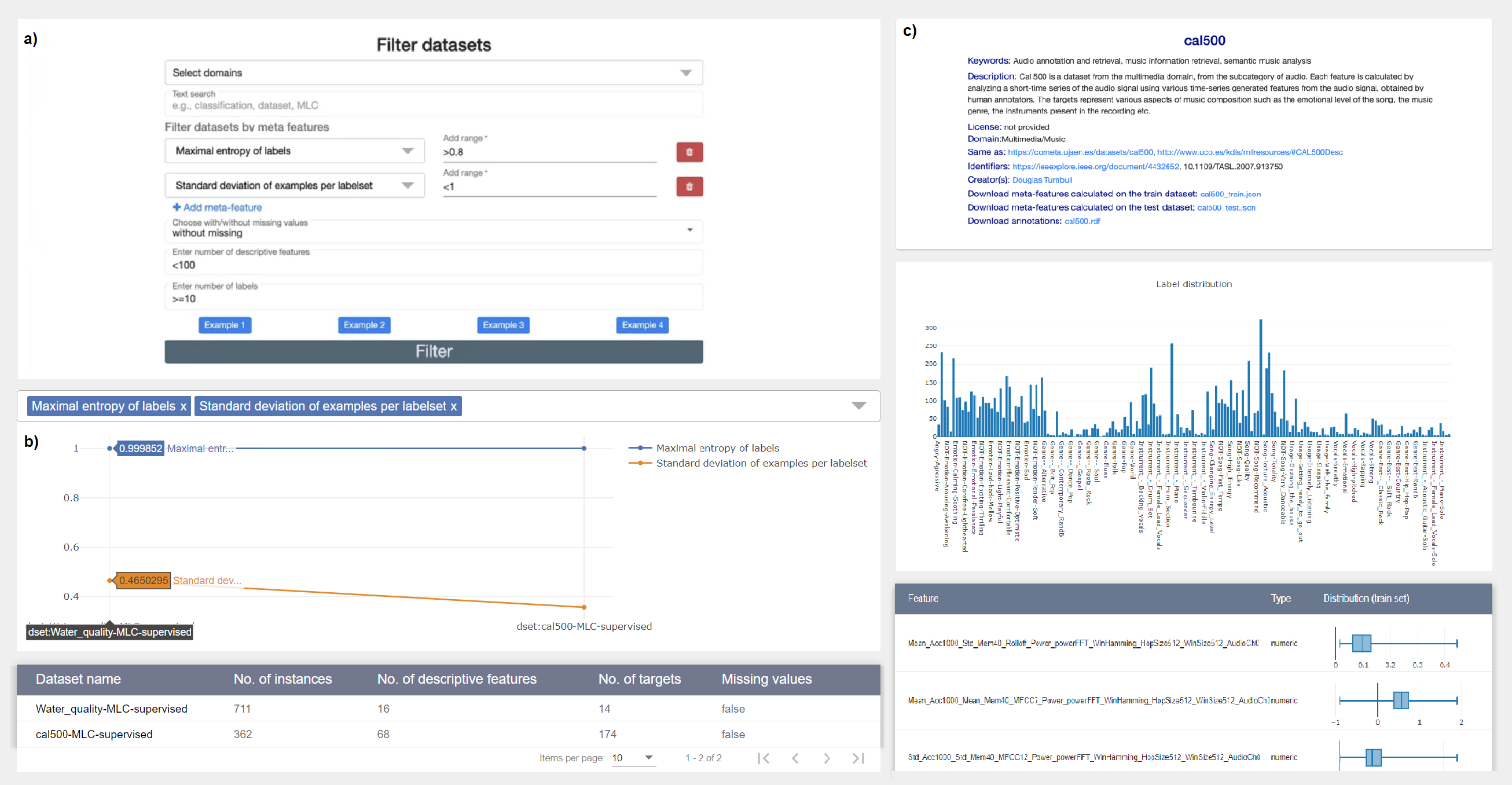}
\caption{A screenshot of the multiple views developed within the GUI of the MLC data catalogue. \textbf{(a)} A filter for querying datasets. \textbf{(b)} Results from the filter query. \textbf{(c)} A view of an individual dataset. 
} 
\label{fig:MLCdatacatalogue}
\end{figure}

The MLC catalogue contains descriptions of 89 MLC datasets in total. Each dataset is annotated with different descriptors. For semantic annotation of datasets, we have designed an ontology-based schema that enables the description of multiple aspects of MLC datasets. The schema is an adaptation of a more general annotation schema that covers a broader range of machine learning tasks presented in Kostovska et al. \cite{kostovska2020semantic}. We can broadly categorize the semantic annotations into two groups: (1) dataset provenance information and (2) machine learning characteristics of the datasets. To allow users to access and interact with the catalogue, we developed a user-friendly web-based system (see Fig.~\ref{fig:MLCdatacatalogue}) to inspect the MLC meta-descriptors. Furthermore, the meta-descriptors are available for cross-comparison with similar datasets present in the catalogue via an interactive visualization engine. 

\begin{figure}[h]
\centering
\includegraphics[width=0.875\textwidth]{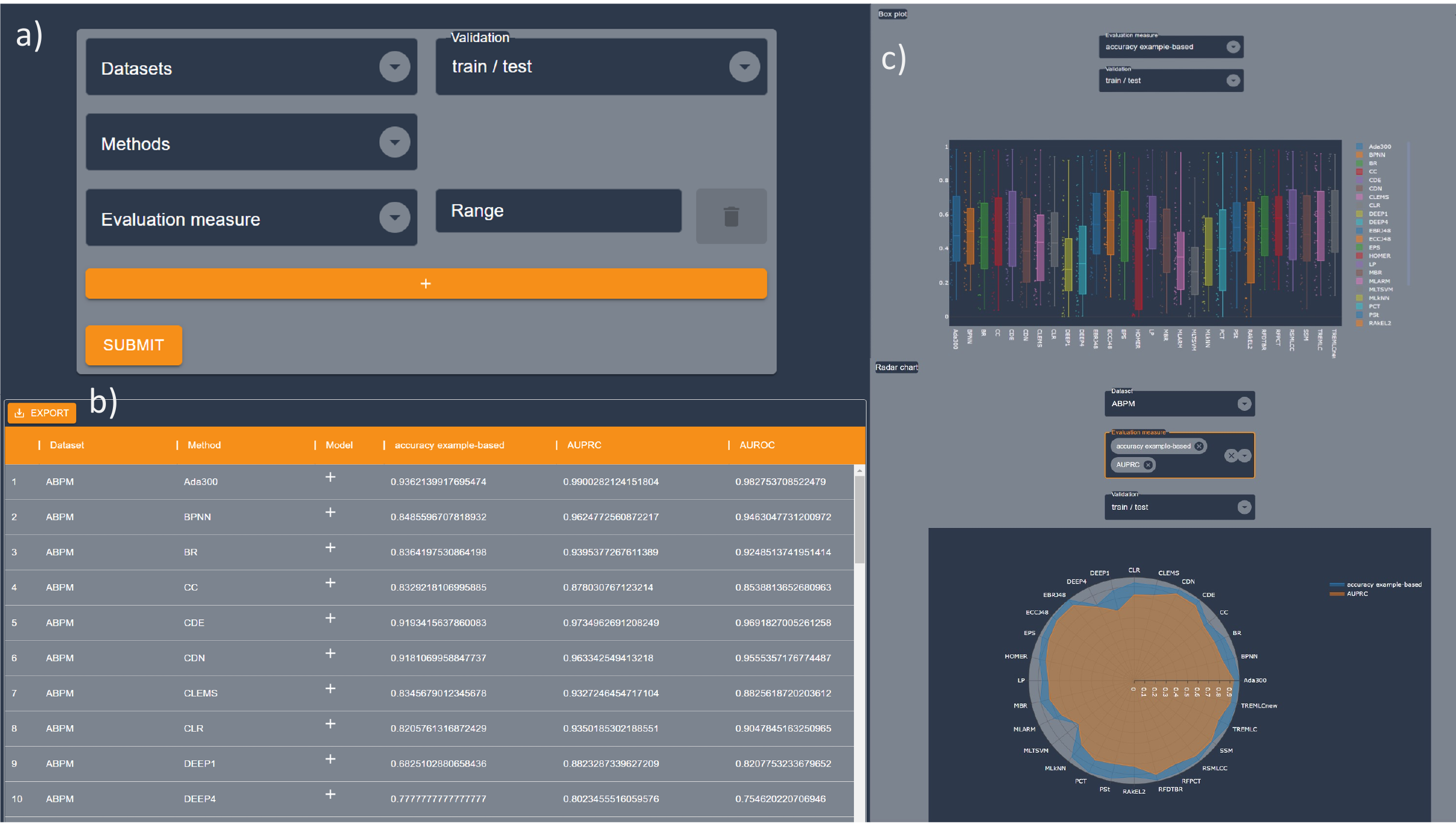}
\caption{A screenshot of the multiple views developed within the GUI of the web-based system. \textbf{(a)} A filter for querying benchmark data. \textbf{(b)} Results from the filter query. \textbf{(c)} Visualization of performance data. } \label{fig:benchmarkGUI}
\end{figure}

The system for MLC benchmark data focuses on the performance data obtained from MLC benchmark studies. The system currently stores MLC benchmark data generated from a comprehensive, comparative study of MLC methods~\cite{bogatinovski2022ESWA1}.  The MLC benchmark data can be divided into 2 categories: (1) data describing the performance landscape of the MLC methods in terms of selected set of evaluation measures; and (2) description of the induced models with their full set of (hyper-)parameters. Apart from performance data, the system relies on semantic annotations that describe the whole predictive modeling pipeline. The semantic annotations are based on a modified version of the semantic scheme presented in \cite{tolovski2020semantic}. Finally, a separate GUI has been developed to easily access the MLC benchmark data. Users can query the data by different criteria (e.g., find all benchmark data related to a given method) and can interactively inspect it via different visualizations (see Fig.~\ref{fig:benchmarkGUI}). 

\section{Discussion and Conclusion}
The key novelty introduced in both systems is that all dataset descriptions are enhanced with semantic annotations (metadata) based on terms from ontologies and controlled vocabularies. The semantic annotations provide the means to develop several useful functionalities of the catalogue: (1) Semantic search over the corpus of annotated datasets; (2) Querying not only the asserted but also the implicitly encoded knowledge in the ontologies by using reasoners; and (3) Improved interoperability of the datasets with external data that follow the same conventions of data representation and management.

From an ML viewpoint, the benefits of the two MLC semantic data resources are threefold. First, the practitioners and non-machine-learning experts can better understand which MLC method to use for their specific use case or system. The use of the proposed resources can reduce the user's learning curve when selecting a method by a non-expert and promote adoption and trust in machine learning across existing and new domains. Second, it can allow experts to jointly reason about the properties of the learning task across different problems. Consequently, it can lead to a better understanding of the task and introduce novel MLC methods that can address the properties of the task under specific conditions. Third, the resources can be used as a benchmark environment to promote transparency when reporting results for a novel MLC method or cross-comparing different results.

\noindent\textbf{Acknowledgment.} The authors would like to acknowledge the support of the Slovenian Research Agency through the project J2-9230, research program P2-0103 and the young researcher grant to Ana Kostovska, as well as of the European Commission through the project TAILOR - Foundations of Trustworthy AI - Integrating Reasoning, Learning and Optimization (grant No. 952215).
%
%
%
\bibliographystyle{splncs04}
\bibliography{mybibliography.bib}
%




\end{document}